\documentclass[runningheads]{llncs}
\usepackage{cite}
\usepackage{amsmath,amssymb,amsfonts}
\usepackage{algorithmic}
\usepackage{subfig}
\usepackage{graphicx,dblfloatfix}
\usepackage{caption}
\usepackage{textcomp}
\usepackage{xcolor}
\usepackage{multirow}
\usepackage[title]{appendix}
\usepackage{enumitem}

\DeclareMathOperator{\E}{\mathbb{E}}

\begin{document}

\title{An Empirical Study on GANs with Margin Cosine Loss and Relativistic Discriminator}

\titlerunning{An Empirical Study on GANs}

\author{Cuong V.  Nguyen\inst{1} \and
Tien-Dung Cao\inst{2} \and
Tram Truong-Huu\inst{3} \and
Khanh N. Pham\inst{1} \and
Binh T. Nguyen\inst{1,4}}
\authorrunning{C. Nguyen et al.}
%
\institute{Entropy Labs\\ 
\email{team@entropyart.io}
\and
School of Engineering, Tan Tao University, Long An, Vietnam \\ 
\email{dung.cao@ttu.edu.vn}
 \and
Institute for Infocomm Research (I$^2$R) \\ Agency for Science, Technology and Research (A*STAR), Singapore \\
\email{tram.truong-huu@ieee.org} 
\and
VNU HCM - University of Science, Vietnam\\
\email{ngtbinh@hcmus.edu.vn}
}

\maketitle

\begin{abstract}
Generative Adversarial Networks (GANs) have emerged as useful generative models, which are capable of implicitly learning data distributions of arbitrarily complex dimensions. However, the training of GANs is empirically well-known for being highly unstable and sensitive. The loss functions of both the discriminator and generator concerning their parameters tend to oscillate wildly during training. Different loss functions have been proposed to stabilize the training and improve the quality of images generated. In this paper, we perform an empirical study on the impact of several loss functions on the performance of standard GAN models, Deep Convolutional Generative Adversarial Networks (DCGANs). We introduce a new improvement that employs a relativistic discriminator to replace the classical deterministic discriminator in DCGANs and implement a margin cosine loss function for both the generator and discriminator. This results in a novel loss function, namely \textit{Relativistic Margin Cosine Loss} (RMCosGAN). We carry out extensive experiments with four datasets: CIFAR-$10$, MNIST, STL-$10$, and CAT. We compare RMCosGAN performance with existing loss functions based on two metrics: Frechet inception distance and inception score. The experimental results show that RMCosGAN outperforms the existing ones and significantly improves the quality of images generated.
\end{abstract}

\section{Introduction}

Generative Adversarial Networks (GANs)~\cite{SGAN2014} have recently become one of the most incredible techniques in machine learning with the capability of implicitly learning data distributions of arbitrarily complex dimensions. To achieve such a capability, a standard GAN model is generally equipped with two adversarial components: a \textit{generator} and a \textit{discriminator} that are neural networks. By Giving a point in a latent space, the generator aims to generate a synthetic sample that is the most similar to a sample in a data distribution of interest. In contrast, the discriminator aims at separating samples produced by the generator from the actual samples in the data distribution. Consequently, the generator and discriminator play a two-player minimax game in which each player has its own objective function, also known as the loss function. 

\subsection{Application of GANs in Natural Image Processing}

Since its emergence, GANs have been applied in various areas of daily lives~\cite{Hong2019_ACM_SURVEY}. In natural image processing, GANs have been used to generate realistic-like images by sampling from latent space. In~\cite{Jetchev_2017}, Jetchev \textit{et al.} proposed a GAN model, namely Conditional Analogy Generative Adversarial Network (CAGAN), for solving image analogy problems, e.g., automatic swapping of clothing on fashion model photos. In~\cite{Zhang_2017}, Zhang \textit{et al.} investigated the text to image synthesis and presented Stacked Generative Adversarial Networks (StackGAN) that can achieve significant improvements in generating photo-realistic images conditioned on text descriptions in comparison with other state-of-the-art techniques. In~\cite{Perarnau2016}, Perarnau \textit{et al.} studied an encoder in a restrictive setting within the GAN framework, namely Invertible Conditional GANs (IcGANs). This framework can help reconstruct or modify real images with deterministic complex modifications and have multiple applications in image editing.

\subsection{Challenges in GAN Training}

Despite significant advantages and widespread application, training of GANs is empirically well-known for being highly unstable and sensitive, leading to a fluctuation of the quality of generated images. The loss functions of both the discriminator and generator concerning their parameters tend to oscillate wildly during training, for theoretical reasons investigated in~\cite{unstablegan:2017}. Among the identified problems, saturation and non-saturation problems have the most significant impact on the performance of GANs. These problems are related to the dynamics of gradient descent algorithms that prevent the generator and discriminator from reaching the optimal values of the training parameters.

The saturation problem is caused by the fact that the discriminator successfully rejects samples created by the generator with high confidence, leading to the gradient vanishing of the generator. It turns out that the generator is not able to learn as quickly as the discriminator, making the model not be trained adequately before the training stops. Modifying the loss function of the generator to a non-saturating one can solve the saturation problem. Still, it raises another problem for the discriminator that could not learn properly to differentiate the actual samples and those generated by the generator. Last but not least, GANs have also been observed to display signs of mode collapse, which occurs when the generator finds only a limited variety of data samples that ``work well'' against the discriminator and repeatedly generates similar copies of data samples.

There exist several works that aim at overcoming the caveats of GANs. Such works include developing different GAN models~\cite{WGAN2017,Sobolev2017}, defining various loss functions~\cite{HingeLimYe2017,HingeTran2017} and designing a relativistic discriminator~\cite{RelativisticGANs2018}. However, these techniques have a partial success of improvement of stability as well as data quality, which are not consistent and significant as most models can reach similar scores with sufficient hyper-parameter optimization and computation~\cite{AreGanEqual2018}.

\subsection{Our Contributions}

In this paper, we carry out an empirical study on the impact of loss functions on the performance of GANs. We develop a novel loss function that improves the quality of generated images. In summary, our contributions are as follows.
\begin{itemize}
  \item We adopt Deep Convolutional Generative Adversarial Networks (DCGANs)~\cite{dcgan:2015} as a baseline architecture where the discriminator is either deterministic or relativistic. With DCGANs, we implement seven loss functions, including cross-entropy (CE), relativistic cross-entropy (R-CE), relativistic average cross-entropy (Ra-CE), least square (LS), relativistic average least square (Ra-LS), hinge (Hinge), and relativistic average hinge (Ra-Hinge).
   
  \item We develop a novel loss function namely \textit{Relativistic Margin Cosine Loss} (RMCosGAN). To the best of our knowledge, this is the first work to improve GAN performance by combining the relativistic discriminator and large margin cosine loss function. 
   
  \item We carry out extensive experiments with four datasets including CIFAR-10~\cite{CIFAR2009}, MNIST~\cite{MNIST1998}, STL-10~\cite{STL2011}, and CAT~\cite{CAT2008}. We compare the performance of RMCosGAN with that of the above seven loss functions with two performance metrics: Frechet inception distance and inception score. We empirically analyze the impact of different parameters and the usefulness of performance metrics in evaluating GAN performance.  
  \item We also open source code and pre-trained models on the GitHub repository\footnote{RMCosGAN GitHub Repository: \url{https://github.com/cuongvn08/RMCosGAN}}. The source code repository provides readers with further details that may not be available in this paper due to the space limit and allow the readers to replicate all experiments presented in this paper. 
\end{itemize}

The rest of the paper is organized as follows. In Section~\ref{sec:method}, we first provide some background knowledge on GANs and then present our proposed loss function. In Section~\ref{sec:experiment}, we present the experiments and analysis of results before we conclude the paper in Section~\ref{sec:conclusion}.

\section{From Cross-entropy to Relativistic Margin Cosine Loss}
\label{sec:method}

\subsection{Standard Generative Adversarial Networks}

Let $\mathcal{X} \subseteq \mathbb{R}^d$ be the \textit{data} space and $\mathcal{Z} \subseteq \mathbb{R}^L$ be the \textit{latent} space, we define $G_{\boldsymbol{\theta}}$ and $D_{\boldsymbol{\psi}}$ to be the generator and discriminator networks of standard GANs (SGAN). The objective function of standard GANs is a saddle point problem defined as 
\begin{equation}
\min\limits_{\boldsymbol{\theta}} \max\limits_{\boldsymbol{\psi}} \mathcal{L}_{\text{SGAN}} (\boldsymbol{\theta}, \boldsymbol{\psi}) 
= - \E\limits_{\boldsymbol{x} \sim{}p_{\mathcal{X}}(\boldsymbol{x})} [\log{}D_{\boldsymbol{\psi}}(\boldsymbol{x})] - \E\limits_{\boldsymbol{z}\sim{}p_{\mathcal{Z}}(\boldsymbol{z})} [\log(1 - D_{\boldsymbol{\psi}}(G_{\boldsymbol{\theta}}(\boldsymbol{z})))] 
\end{equation}
The generator and discriminator mappings depend on the trainable parameters $\boldsymbol{\theta}$ and $\boldsymbol{\psi}$. In standard GANs, the choice of parameterization for both of these mappings is given by the use of fully connected neural networks trained using gradient descent via backpropagation. Given a prior distribution $p_{\mathcal{Z}}(\boldsymbol{z})$ of the latent variable usually chosen to be a standard Gaussian, a perfect $D_{\boldsymbol{\psi}}$ takes in an input $\boldsymbol{x} \in \mathcal{X}$ and outputs a probability $y$ where $y=0$ if $\boldsymbol{x}$ is a generated sample, i.e., $\boldsymbol{\tilde{x}} := G_{\boldsymbol{\theta}}(\boldsymbol{z})$ for $\boldsymbol{z} \sim p_{\mathcal{Z}}(\boldsymbol{z})$ and $y=1$ if $\boldsymbol{x}$ is a real sample, i.e., $\boldsymbol{x} \sim p_{\mathcal{X}}(\boldsymbol{x})$. In standard GANs, the discriminator $D_{\boldsymbol{\psi}}$ is defined as $D_{\boldsymbol{\psi}} := \sigma (C_{\boldsymbol{\psi}})$ where $\sigma$ is the sigmoid activation function and $C_{\boldsymbol{\psi}}$ is a real-valued function known as the \textit{critic}, which represents the pre-activation logits of the discriminator network. Based on standard GANs, Radford \textit{et al.} developed DCGANs~\cite{dcgan:2015} in which the generator and discriminator are both convolutional neural networks. Specifically, the generator is an upsampling CNN made up of transposed convolutional layers, while the discriminator is a typical downsampling CNN with convolutional layers.

\subsection{Relativistic Discriminator}

To address the shortcomings of standard GANs, Jolicoeur-Martineau~\cite{RelativisticGANs2018} proposed to use a relativistic discriminator for a non-saturating loss and empirically shown that a relativistic discriminator can ensure divergence minimization and produce sensible output. The objective function of a standard GAN using a relativistic discriminator (RSGAN) is a saddle point problem defined as follows:

\begin{align} 
\!\!\!\!\min\limits_{\boldsymbol{\theta}} \max\limits_{\boldsymbol{\psi}} \mathcal{L}_{\text{RSGAN}} (\boldsymbol{\theta}, \boldsymbol{\psi}) & = 
- \E\limits_{\boldsymbol{x} \sim p_{\mathcal{X}}(\boldsymbol{x}), \boldsymbol{z} \sim p_{\mathcal{Z}}(\boldsymbol{z})} [\log (\sigma ( C_{\boldsymbol{\psi}}(\boldsymbol{x}) - C_{\boldsymbol{\psi}}(G_{\boldsymbol{\theta}}(\boldsymbol{z}))))] \nonumber \\
&- \E\limits_{\boldsymbol{x} \sim p_{\mathcal{X}}(\boldsymbol{x}), \boldsymbol{z} \sim p_{\mathcal{Z}}(\boldsymbol{z})} [\log (\sigma ( C_{\boldsymbol{\psi}}(G_{\boldsymbol{\theta}}(\boldsymbol{z})) - C_{\boldsymbol{\psi}}(\boldsymbol{x})))]
\label{eq:RSGAN}
\end{align}

Rather than determining whether sample $\boldsymbol{x}$ is a real sample ($y=1$) or a generated sample ($y=0$), the relativistic discriminator estimates the probability that a given sample in the dataset is more realistic than a randomly sampled generated data. 

 \subsection{Relativistic Margin Cosine Loss}

\begin{figure}[t]
    \centering
    \includegraphics[width=\textwidth]{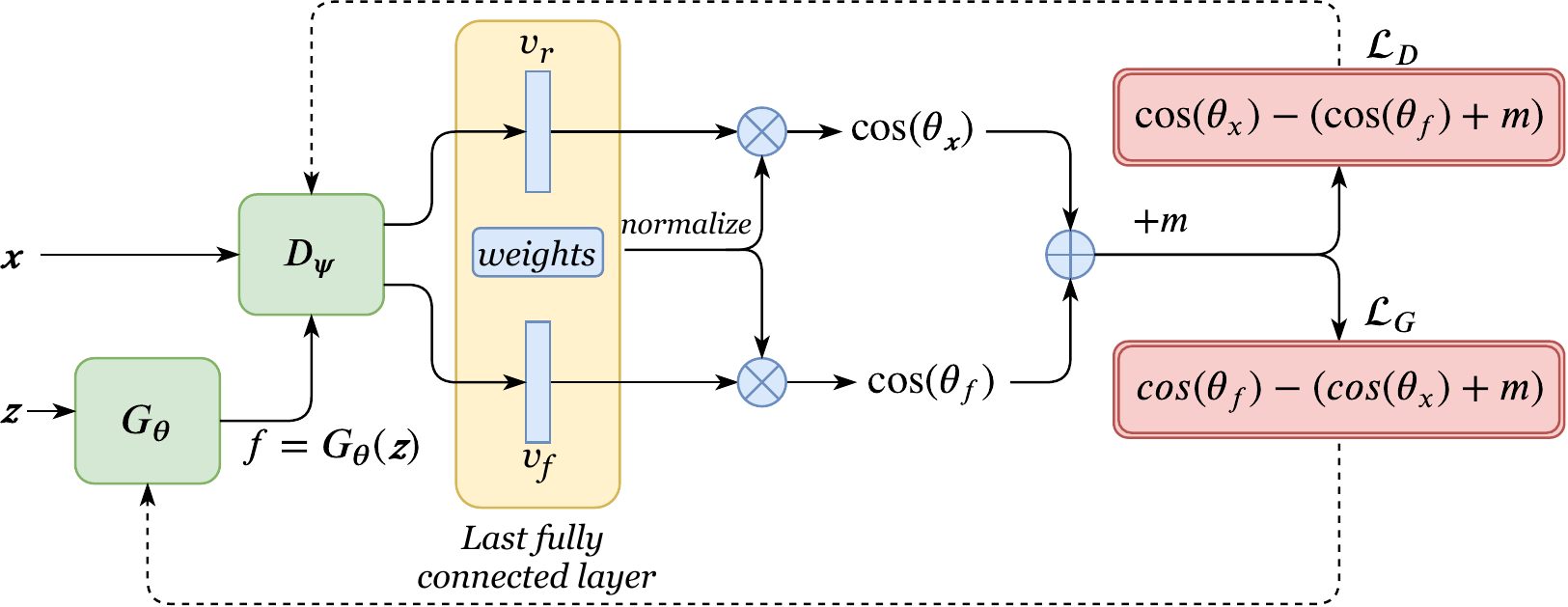}
    \caption{Architecture of Relativistic Margin Cosine GANs.}
    \label{fig:architecture}
    \vspace{-2.5ex}
\end{figure}

The margin cosine loss function presented in~\cite{CosFace2018} serves to maximize the degree of inter-class variance and to minimize intra-class variance in discriminating between real and generated samples.  We incorporate the margin cosine loss function in place of the binary cross-entropy loss in the RSGAN model, resulting in a novel loss function, namely Relativistic Margin Cosine Loss (RMCosGAN). We name the GAN model using RMCosGAN loss function with a relativistic discriminator as Relativistic Margin Cosine GAN (RMCosGAN). 

As depicted in Fig.~\ref{fig:architecture}, we re-define the critic $C_{\boldsymbol{\psi}}$ that represents the pre-activation logits of both the generator and discriminator networks. Let $W$ be the weight vector of the last fully connected layer $L$. The incoming activation vector produced by layer $L-1$ of a real data sample $\boldsymbol{x}$ is $v_r$ and that of a fake sample $G_{\boldsymbol{\theta}}(\boldsymbol{z})$ generated from latent point $\boldsymbol{z}$ is $v_f$. Conventionally, the pre-activation logits of real sample $\boldsymbol{x}$ at layer $L$ with a zeroed bias is computed as $C_{\boldsymbol{\psi}}(\boldsymbol{x}):= W^Tv_r = ||W|||v_r||\cos(\theta_{\boldsymbol{x}})$ where $\theta_{\boldsymbol{x}}$ is  the angle between $W$ and $v_r$. Similarly, that of a fake sample $G_{\boldsymbol{\theta}}(\boldsymbol{z})$ is $C_{\boldsymbol{\psi}}(G_{\boldsymbol{\theta}}(\boldsymbol{z})):= W^Tv_f = ||W|||v_r||\cos(\theta_{f})$ where $\theta_{f}$ is  the angle between $W$ and $v_f$. To enable effective feature learning, the norm of $W$ should be invariable. Thus, we fix $||W|| = 1$ by applying an $L_2$ normalization. Furthermore, as the norm of feature vectors ($||v_r||$ and $||v_f||$) does not contribute to the cosine similarity between the two feature vectors, we set $||v_r|| = ||v_f||$ to a constant $(s)$. Consequently, the pre-activation logits solely depend on the cosine of the angle. This suggests us defining the critic as $C_{\boldsymbol{\psi}}(\boldsymbol{x}):= \cos(\theta_{\boldsymbol{x}})$  and  $C_{\boldsymbol{\psi}}(G_{\boldsymbol{\theta}}(\boldsymbol{z})):= \cos(\theta_f)$  and where $\theta_x$ and $\theta_f$ are defined as above. The fixed parameter $m \geqslant 0$ is also introduced to control the magnitude of the cosine margin. We end up defining the loss functions for the discriminator ($\mathcal{L}_D$) and that of the generator ($\mathcal{L}_G$) as shown in Fig.~\ref{fig:architecture}. The objective function of RMCosGAN is a saddle point problem finally defined as follows: 
\begin{align} 
&\min\limits_{\boldsymbol{\theta}} \max\limits_{\boldsymbol{\psi}} \mathcal{L}_{\text{RMCosGAN}} (\boldsymbol{\theta},\boldsymbol{\psi}) = \nonumber\\ 
&\quad\quad\quad\quad\quad\quad\quad - \E\limits_{\boldsymbol{x} \sim p_{\mathcal{X}}(\boldsymbol{x}), \boldsymbol{z} \sim p_{\mathcal{Z}}(\boldsymbol{z})}[\log (\sigma (s(C_{\boldsymbol{\psi}}(\boldsymbol{x}) -  (C_{\boldsymbol{\psi}}(G_{\boldsymbol{\theta}}(\boldsymbol{z}))+ m)))] - \nonumber\\
 &\quad\quad\quad\quad\quad\quad\quad\E\limits_{\boldsymbol{x} \sim p_{\mathcal{X}}(\boldsymbol{x}), \boldsymbol{z} \sim p_{\mathcal{Z}}(\boldsymbol{z})}[\log (\sigma (s( C_{\boldsymbol{\psi}}(G_{\boldsymbol{\theta}}(\boldsymbol{z})) -  (C_{\boldsymbol{\psi}}(\boldsymbol{x})+ m)))]
\end{align}
 We note that while the relativistic discriminator and margin cosine loss function have been developed independently, to the best of our knowledge, our work is the first to combine them in an integrated model to further improve the quality of the samples generated. 

\subsection{Further Analysis}

In this section, we aim to analyze the properties of the objective function $\mathcal{L}_{\text{RMCosGAN}} (\boldsymbol{\theta},\boldsymbol{\psi})$. First, the derivative of this objective function with respect to the parameter $m$ can be computed as follows:
\begin{eqnarray}
&& \frac{\partial}{\partial m} \left[{\mathcal{L}_{\text{RMCosGAN}} (\boldsymbol{\theta},\boldsymbol{\psi})}\right] = - \frac{\partial}{\partial m} \left\{{ \E\limits_{\boldsymbol{x} \sim p_{\mathcal{X}}(\boldsymbol{x}), \boldsymbol{z} \sim p_{\mathcal{Z}}(\boldsymbol{z})}[\log (\sigma (s(C_{\boldsymbol{\psi}}(\boldsymbol{x}) -  (C_{\boldsymbol{\psi}}(G_{\boldsymbol{\theta}}(\boldsymbol{z}))+ m)))]
}\right\}\nonumber \\
&-& \frac{\partial}{\partial m} \left\{{ \E\limits_{\boldsymbol{x} \sim p_{\mathcal{X}}(\boldsymbol{x}), \boldsymbol{z} \sim p_{\mathcal{Z}}(\boldsymbol{z})}[\log (\sigma (s( C_{\boldsymbol{\psi}}(G_{\boldsymbol{\theta}}(\boldsymbol{z})) -  (C_{\boldsymbol{\psi}}(\boldsymbol{x})+ m)))]}\right\}
\end{eqnarray}
Now, we consider
\begin{equation}
 h(\theta,\psi)=C_{\boldsymbol{\psi}}(\boldsymbol{x}) -  C_{\boldsymbol{\psi}}(G_{\boldsymbol{\theta}}(\boldsymbol{z})), 
\end{equation}
then
\begin{eqnarray}
    & &\frac{\partial}{\partial m} \left\{{ \E\limits_{\boldsymbol{x} \sim p_{\mathcal{X}}(\boldsymbol{x}), \boldsymbol{z} \sim p_{\mathcal{Z}}(\boldsymbol{z})}[\log (\sigma (s(C_{\boldsymbol{\psi}}(\boldsymbol{x}) -  (C_{\boldsymbol{\psi}}(G_{\boldsymbol{\theta}}(\boldsymbol{z}))+ m)))]
}\right\} \nonumber \\
&=& \frac{\partial}{\partial m} \left[{ \E\limits_{\boldsymbol{x} \sim p_{\mathcal{X}}(\boldsymbol{x}), \boldsymbol{z} \sim p_{\mathcal{Z}}(\boldsymbol{z})}[\log (\sigma (s(h(\theta,\psi) - m)))]
}\right] \nonumber \\
&=&  \E\limits_{\boldsymbol{x} \sim p_{\mathcal{X}}(\boldsymbol{x}), \boldsymbol{z} \sim p_{\mathcal{Z}}(\boldsymbol{z})} \left[{\frac{\partial}{\partial m} \left\{{\log (\sigma (s(h(\theta,\psi) - m)))}\right\}}\right] \nonumber \\
&=&  \E\limits_{\boldsymbol{x} \sim p_{\mathcal{X}}(\boldsymbol{x}), \boldsymbol{z} \sim p_{\mathcal{Z}}(\boldsymbol{z})} \left[{
\frac{1}{\sigma (s(h(\theta,\psi) - m))}\times \frac{\exp\{s(h(\theta,\psi) - m)\}}{\left[{\exp\{s(h(\theta,\psi) - m)\}}\right]^2} \times (-s)
}\right] \nonumber \\
&=&  \E\limits_{\boldsymbol{x} \sim p_{\mathcal{X}}(\boldsymbol{x}), \boldsymbol{z} \sim p_{\mathcal{Z}}(\boldsymbol{z})} \left[{
\frac{-s}{\sigma (s(h(\theta,\psi) - m))}\times \frac{\exp\{s(h(\theta,\psi) - m)\}}{\left[{\exp\{s(h(\theta,\psi) - m)\}}\right]^2}
}\right] \le 0
\end{eqnarray}
Similarly, we can also prove that
\begin{eqnarray}
    \frac{\partial}{\partial m} \left\{{ \E\limits_{\boldsymbol{x} \sim p_{\mathcal{X}}(\boldsymbol{x}), \boldsymbol{z} \sim p_{\mathcal{Z}}(\boldsymbol{z})}[\log (\sigma (s( C_{\boldsymbol{\psi}}(G_{\boldsymbol{\theta}}(\boldsymbol{z})) -  (C_{\boldsymbol{\psi}}(\boldsymbol{x})+ m)))]}\right\} \le 0
\end{eqnarray}
It turns out that 
\begin{eqnarray}
\frac{\partial}{\partial m} \left[{\mathcal{L}_{\text{RMCosGAN}} (\boldsymbol{\theta},\boldsymbol{\psi})}\right] \ge 0
\end{eqnarray}
for all $\theta$ and $\psi$.\\
It is worth noting that$
    \mathcal{L}_{\text{RaSGAN}} (\boldsymbol{\theta}, \boldsymbol{\psi}) = \mathcal{L}_{\text{RMCosGAN}} (\boldsymbol{\theta},\boldsymbol{\psi}) |_{m = 0}
$.
Thus, we can obtain the following theorem:
\begin{theorem}
The objective function $\mathcal{L}_{\text{RMCosGAN}} (\boldsymbol{\theta},\boldsymbol{\psi})$ is a monotonically decreasing function with respect to $m$. Especially, we have:
\begin{enumerate}[label=(\alph*)]
    \item $\mathcal{L}_{\text{RSGAN}} (\boldsymbol{\theta}, \boldsymbol{\psi})> \mathcal{L}_{\text{RMCosGAN}} (\boldsymbol{\theta},\boldsymbol{\psi})$, when $m<0$.
    \item  $\mathcal{L}_{\text{RSGAN}} (\boldsymbol{\theta}, \boldsymbol{\psi}) < \mathcal{L}_{\text{RMCosGAN}} (\boldsymbol{\theta},\boldsymbol{\psi})$, when $m>0$.
    \item $\mathcal{L}_{\text{RSGAN}} (\boldsymbol{\theta}, \boldsymbol{\psi}) = \mathcal{L}_{\text{RMCosGAN}} (\boldsymbol{\theta},\boldsymbol{\psi})$, when $m=0$.
\end{enumerate}
The results still hold for other activation functions having non-negative derivatives, such as ReLu, Tanh, SoftPlus, Arctan. 
\end{theorem}

\section{Experiments}
\label{sec:experiment}
\subsection{Experimental Settings}

\subsubsection{Implementation Details.} 
We implemented the models in PyTorch and trained them using Adam optimizer. All the experiments were run on two workstations:
\begin{itemize}
    \item An Intel(R) Core(TM) i$7$ with $2$ CPUs @ $2.4$GHz,  $128$GB of RAM and an Nvidia GeForce RTX $2080$Ti GPU of $11$GB of memory.
    \item A customized desktop with AMD Ryzen Threadripper $2950$X $16$-core processor @ $3.5$GHz, $64$ GB of RAM and $2$ Nvidia GeForce RTX $2080$Ti GPUs, each having $11$ GB of memory.      
\end{itemize}

\subsubsection{Datasets and Preprocessing.} We used four benchmark datasets of CIFAR-$10$~\cite{CIFAR2009}, MNIST~\cite{MNIST1998}, STL-$10$~\cite{STL2011} and CAT~\cite{CAT2008}. There are $50$K $32\times{}32$ images in CIFAR-$10$, $60$K $28\times{}28$ images in MNIST, $5$K $96\times{}96$ images in STL-$10$ and $9408$ $64\times{}64$ images in the CAT dataset. We resized the original images of MNIST to $32\times{}32$ to be able to use the same architecture with the CIFAR-$10$ dataset and converted them to RGB images to compute FID and IS. We cropped the faces of the cats in the CAT dataset according to the annotation. We resized the original images of STL-$10$ to $48\times{}48$ as well. All of the images in the datasets were normalized at a mean of $0.5$ and a standard deviation of $0.5$. 

\subsubsection{Evaluation Metrics.} We used the Frechet inception distance (FID)~\cite{FID2017} and inception score (IS)~\cite{IS2016} for quantitative evaluation. Both FID and IS are measured using a pre-trained inception model\footnote{\url{http://download.tensorflow.org/models/image/imagenet/inception-2015-12-05.tgz}}. FID calculates the Wasserstein-2 distance between a generated image and a real image in the feature space of an inception network. Lower FID indicates a closer distance between the real image and generated image, indicating a better image quality. For MNIST, STL-10 and CAT, FID is measured based on the generated images and real images in the training dataset. For the CIFAR-10 dataset, FID is measured between the generated images and the pre-calculated FID stats. IS measures the quality and diversity of generated images. The higher IS, the better the diversity of generated images.

\subsubsection{Comparison.} 
We implemented seven loss functions including cross-entropy (CE), relativistic cross-entropy  (R-CE), relativistic average cross-entropy (Ra-CE), least square (LS), relativistic average least square (Ra-LS), hinge  (Hinge), and relativistic average hinge (Ra-Hinge). We compared the performance of RMCosGAN against that of the above seven loss functions. 

\subsubsection{Model Architectures and Hyper-parameters.} The model architectures used in our work are similar to~\cite{RelativisticGANs2018}, which are based on DCGANs~\cite{dcgan:2015}. Two model architectures were used to investigate the performance of all loss functions, the first one is used to evaluate the performance on the datasets of CIFAR-$10$, MNIST, STL-$10$ and the other is used to evaluate the performance on the CAT dataset. Similar to \cite{SpectralNorm2018}, we used spectral normalization in the discriminator and batch normalization in the generator. For further details of the model architectures and training parameters, we refer the readers to our GitHub repository. 

\subsection{Analysis of Results}
\subsubsection{Replication of Recent Work.}
\begin{table}[t]
\centering
\caption{FIDs on CIFAR-$10$ and CAT datasets after 100K training epochs}
\label{tab:replication}
\begin{tabular}{l@{\hskip 0.2in}r@{\hskip 0.2in} r@{\hskip 0.2in} r@{\hskip 0.2in} r@{\hskip 0.2in} } 
\hline
\multirow{2}{*}{\textbf{Loss Function}} & \multicolumn{2}{c}{\textbf{CIFAR-$\mathbf{10}$}} & \multicolumn{2}{c}{\textbf{CAT}} \\
& \textbf{\cite{RelativisticGANs2018}} & \textbf{Our work} & \textbf{\cite{RelativisticGANs2018}} & \textbf{Our work}\\
\hline\hline
CE & 40.64 & 33.82 & 16.56 &  11.91\\ 
\hline
R-CE & 36.61 & 40.02 & 19.03 & 10.80 \\
\hline
Ra-CE  & 31.98 & 36.82 & 15.38 & $\mathbf{9.95}$\\ 
\hline
LS  & $\mathbf{29.53}$  & 31.25 & 20.27 & 18.49\\
\hline
Ra-LS  & 30.92 & $\mathbf{30.61}$ & \textbf{11.97} & 13.85 \\ 
\hline
Hinge  & 49.53 & 40.16 & 17.60 & 12.27 \\
\hline
Ra-Hinge  & 39.12  & 42.32 & 14.62 & 10.57\\
\hline
\end{tabular}
\end{table}

We replicated the work presented in~\cite{RelativisticGANs2018}. As shown in Table~\ref{tab:replication}, we observed that there is a discrepancy in the results we obtained and that reported in~\cite{RelativisticGANs2018}. While~\cite{RelativisticGANs2018} reported that LS is the best loss function for CIFAR-$10$ and Ra-LS is the best loss function for CAT, our results show that Ra-LS is the best loss function for CIFAR-$10$ and Ra-CE is the best loss function for CAT. Furthermore, there is no common behavior of the loss functions throughout this experiment. This confirms the instability of GANs and shows that existing loss functions still cannot significantly stabilize GAN training and provide a better quality of images.  

\subsubsection{Performance of RMCosGAN}
We evaluate the performance of RMCosGAN in comparison with the existing ones. In Table~\ref{tab:FID}, we present FID obtained with all the loss functions on the four datasets. The results show that RMCosGAN outperforms the existing loss functions for most of the datasets. For CIFAR-$10$, while the best loss function (Ra-LS) has an FID of $30.61$, RMCosGAN approximates this performance with an FID of $31.34$. It is worth mentioning that RMCosGAN significantly reduces FID on MNIST and STL-$10$. Compared to the second-best loss functions (Ra-CE for MNIST and Ra-Hinge for STL-10), RMCosGAN reduces FID by $21\%$ and $7.5\%$, respectively. Compared to Ra-LS, RMCosGAN has a much lower FID.

We also observed that the relativistic loss functions (R-CE, Ra-CE, Ra-LS and Ra-Hinge) perform much better than their original versions (CE, LS, Hinge) on MNIST and CAT. However, this is not the case when considering CIFAR-10 and STL-10, both relativistic and non-relativistic loss functions behave arbitrarily. This demonstrates that using a relativistic discriminator alone may not be generalized for any datasets. Given such instability of the existing loss functions, the obtained results with RMCosGAN demonstrate its effectiveness over different datasets with diverse data distributions.

\begin{table}[t]
\centering
\caption{FID achieved with all the  datasets at 100K training epochs}
\label{tab:FID}
\begin{tabular}{l@{\hskip 0.2in} r@{\hskip 0.2in} r@{\hskip 0.2in} r@{\hskip 0.2in} r}
\hline
\textbf{Loss Function} & \textbf{CIFAR-10} & \textbf{MNIST} & \textbf{STL-10} & \textbf{CAT}\\
\hline\hline
CE & $33.82\pm0.03$ & $31.87\pm0.14$ & $56.46\pm0.01$ & $11.91\pm0.09$\\
\hline
R-CE & $40.02\pm0.16$ & $18.54\pm0.30$ & $58.53\pm0.28$ & $10.80\pm0.05$\\ 
\hline
Ra-CE & $36.82\pm0.19$ & $16.65\pm0.20$ & $61.87\pm0.19$ & $9.95\pm0.10$\\ 
\hline
LS & $31.25\pm0.23$ & $32.42\pm0.01$ & $57.47\pm0.30$ &  $18.49\pm0.20$\\ 
\hline
Ra-LS & $\mathbf{30.61}\pm\mathbf{0.07}$ & $21.73\pm0.40$ & $59.68\pm0.38$ & $13.85\pm0.10$\\ 
\hline
Hinge & $40.16\pm0.10$ & $20.83\pm0.18$ & $67.43\pm0.09$ & $12.27\pm0.14$\\ 
\hline
Ra-Hinge & $42.32\pm0.09$ & $16.76\pm0.21$ & $56.38\pm0.39$ & $10.57\pm0.17$\\ 
\hline
\textbf{RMCosGAN} & $31.34\pm0.13$ & $\mathbf{13.17}\pm\mathbf{0.05}$ & $\mathbf{52.16}\pm\mathbf{0.35}$ &  $\mathbf{9.48}\pm\mathbf{0.11}$\\ 
\hline
\end{tabular}
\end{table}

In Table~\ref{tab:IS}, we present the performance of the loss functions in terms of IS. We observed that the proposed loss function RMCosGAN performs slightly better than the existing loss functions on CIFAR-10 and STL-10. However, RMCosGAN does not perform well on MNIST and CAT. We carried out an ablation study and realized that IS is not a meaningful performance metric for evaluating GANs. We present further details of this study in the next section.

\begin{table}[t]
    \centering
    \caption{IS achieved with all the datasets at 100K training epochs}
    \label{tab:IS}
    \begin{tabular}{l@{\hskip 0.2in} r@{\hskip 0.2in} r@{\hskip 0.2in} r@{\hskip 0.2in} r}
         \hline
\textbf{Loss Function} & \textbf{CIFAR-10} & \textbf{MNIST} & \textbf{STL-10} & \textbf{CAT}\\
\hline\hline
CE & $6.78\pm0.05$ & $2.30\pm0.02$ & $6.83\pm0.03$ & $3.22\pm0.01$\\ 
\hline
R-CE & $6.49 \pm 0.06$ & $2.41 \pm 0.01$ & $6.26\pm0.03$ & $3.53\pm0.01$\\ 
\hline
Ra-CE & $6.60 \pm 0.02$ & $2.25 \pm 0.01$ & $6.17 \pm 0.06$ & $3.46 \pm 0.02$\\ 
\hline
LS & $6.77 \pm 0.03$ & $2.46 \pm 0.01$ & $6.43 \pm 0.05$ & $\mathbf{4.18} \pm \mathbf{0.01}$\\
\hline
Ra-LS & $6.79 \pm 0.07$ & $\mathbf{2.59} \pm \mathbf{0.01}$ & $6.44 \pm 0.03$ & $3.85 \pm 0.01$\\ 
\hline
Hinge & $6.64\pm0.06$ & $2.51\pm0.02$ & $5.85\pm0.03$ & $3.79\pm0.01$\\ 
\hline
Ra-Hinge & $6.42\pm0.02$ & $2.31\pm0.01$ & $6.28\pm0.07$ & $3.50\pm0.01$\\ 
\hline
\textbf{RMCosGAN} & $\mathbf{7.08}\pm\mathbf{0.06}$ & $2.18\pm0.02$ & $\mathbf{6.86}\pm\mathbf{0.03}$ & $3.58\pm 0.02$ \\ 
\hline
\end{tabular}
\end{table}

\subsection{Ablation Study}
\subsubsection{FID is better than IS in evaluating GANs.}
\begin{figure}[t]
    \centering
    \includegraphics[width=\textwidth]{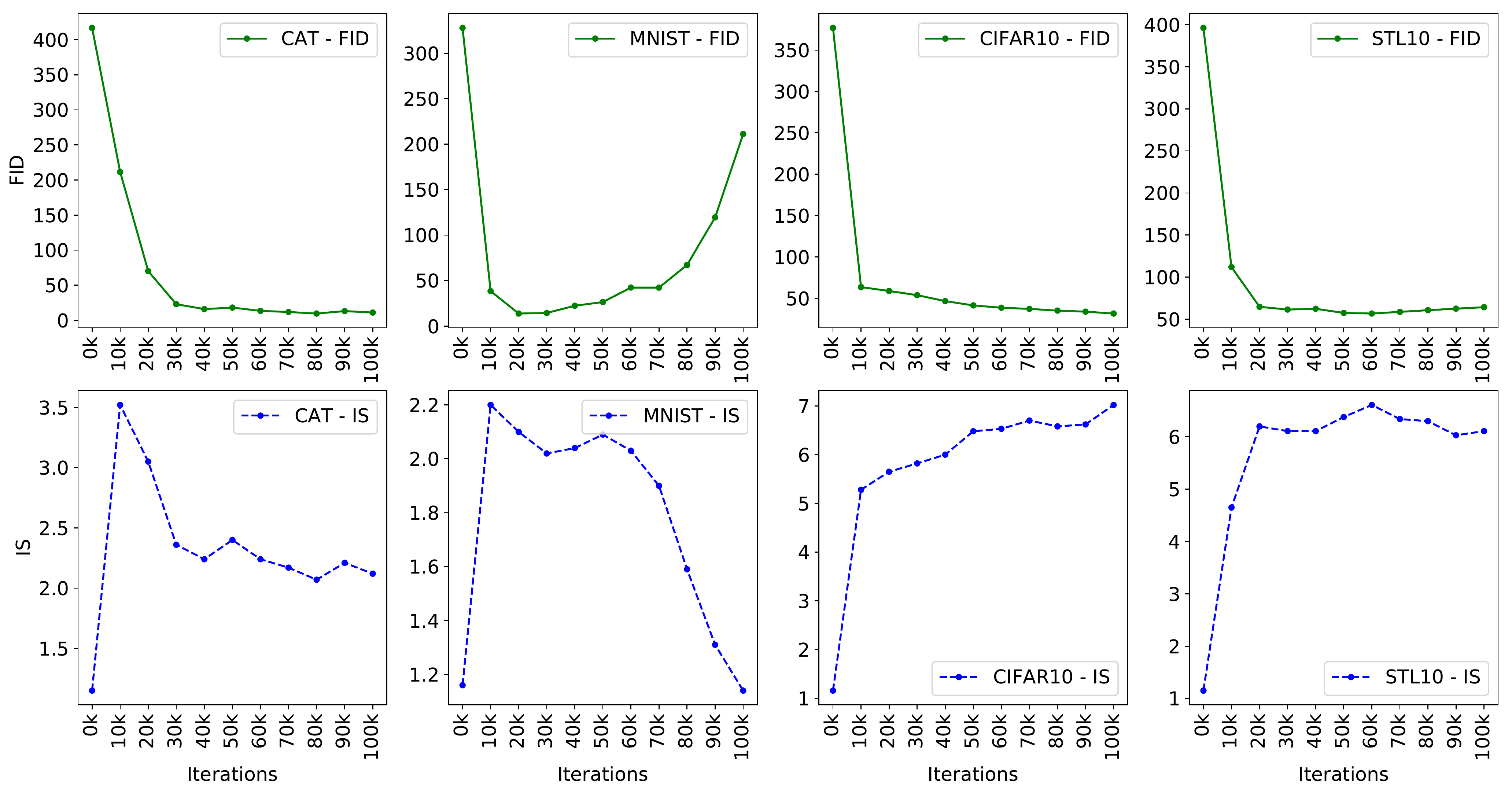}
    \caption{FID and IS at different training epochs on the four datasets.}
    \label{fig:FID_vs_IS}
\end{figure}

\begin{figure}[t]
    \centering
    \includegraphics[width=1.0\textwidth]{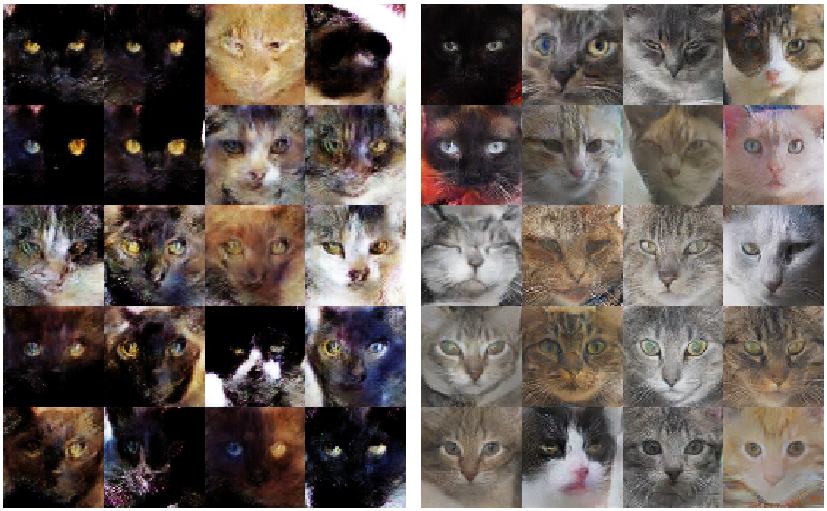}
    \caption{Generated images by GANs trained on CAT dataset. (Left) 20 randomly-generated images at the 15K-th epoch with FID=$112.66$ and IS=3.54. (Right) 20 randomly-generated images at 95K-th epoch with FID$=9.19$ and IS$=2.05$.}
    \label{fig:cat_15k_95}
\end{figure}

In this ablation study, we evaluate the effectiveness and correlation between FID and IS in evaluating GANs. During the training of GANs using RMCosGAN, we measured FID and IS at different training epochs for all the datasets. We plot the experimental results in Fig.~\ref{fig:FID_vs_IS}. We observed that FID and IS of GANs behave similarly when training on CIFAR-10 and STL-10. The longer the training process, the better the values of FID and IS. In other words, FID decreases, whereas IS increases concerning training time. However, this behavior was not consistent when training on CAT and MNIST. While FID keeps decreasing along with the training on the CAT dataset, IS quickly increases and then decreases gradually. The worst case is MNIST, on which FID and IS form two opposite bell curves such that the optimal values of FID and IS for MNIST can be obtained in the middle of training rather than at the end of the training. This also means that a model that achieves a high FID may produce very low IS and vice versa. One of the reasons that explain this behavior is that the distribution of generated data and the pre-trained distribution used for computing FID and IS are too far~\cite{NoteOnInceptionScore}. Thus, a high-quality image can be generated by a model with a low IS. The generated image does not exhibit its diversity to be considered a distinct class. In Fig.~\ref{fig:cat_15k_95}, we present several images that are generated at different moments during the training process to illustrate this behavior. The images on the right side have better quality with a low IS. 
  
\subsubsection{Impact of margin $m$ on RMCosGAN performance.} 
\begin{figure}[t]
\centering
  \subfloat[FID with different margin settings achieved with RMCosGAN on CIFAR-10 and MNIST.]
    {\label{fig:impact_m}\includegraphics[width=0.35\textwidth]{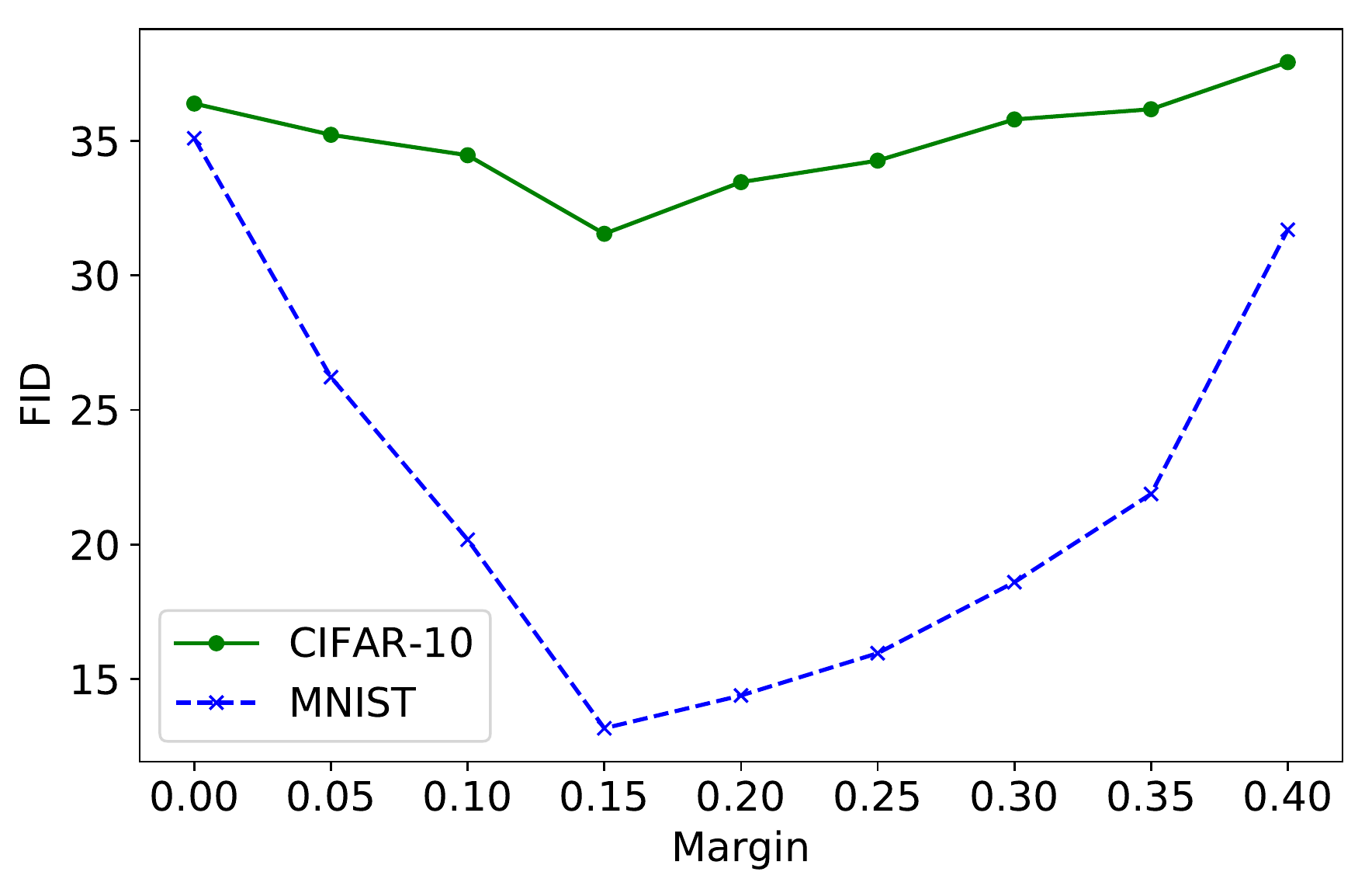}}\hspace{2mm}
  \subfloat[FID with different number of images generated on CIFAR-10.]
    {\label{fig:impact_cifar10}\includegraphics[width=0.29\textwidth]{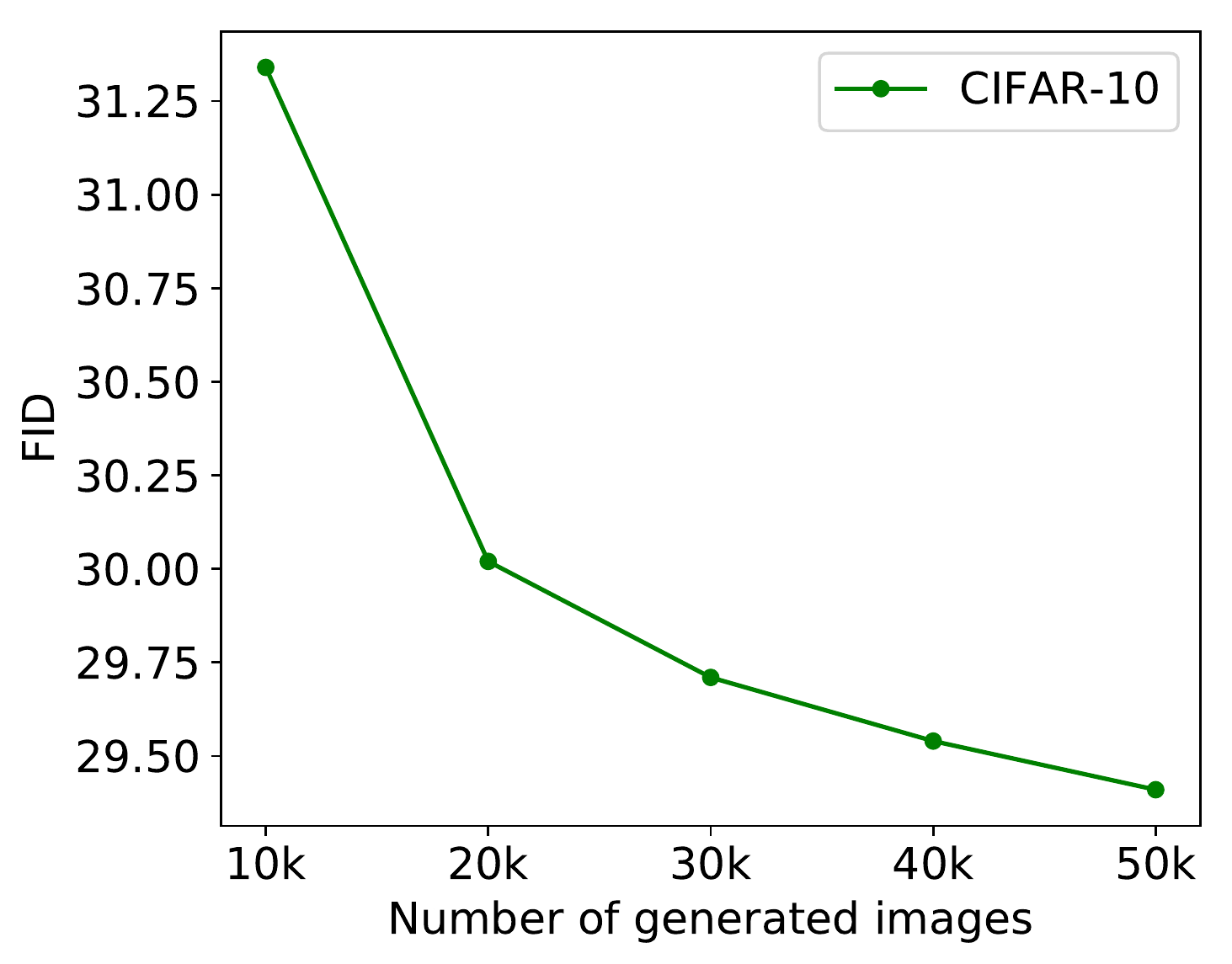}} 
    \hspace{2mm}
  \subfloat[FID with different number of images generated on MNIST.]
    {\label{fig:impact_mnist}\includegraphics[width=0.29\textwidth]{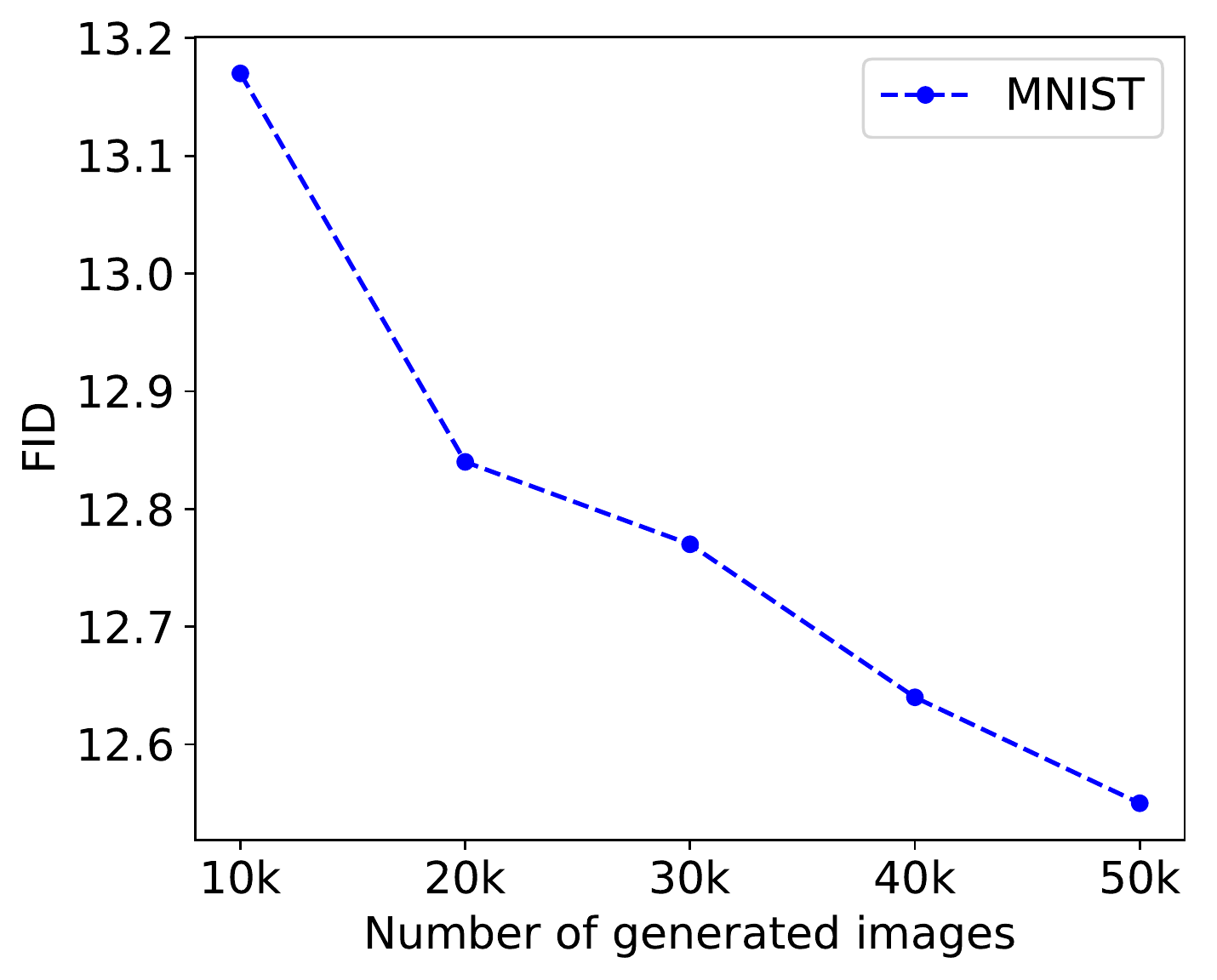}}
\caption{Impact of margin coefficient and the number of generated images on FID.}
\label{fig:impact}
\end{figure}

To explore the impact of cosine margin ($m$) on the performance of RMCosGAN, we set the scale $s=10$ and trained our RMCosGAN model with different margins in the range $[0, 1]$ on two datasets: CIFAR-10 and MNIST. The results are shown in Fig.~\ref{fig:impact_m}. During the training, we observed that RMCosGAN without margin ($m=0$) or with a very large margin ($m>0.4$) performed badly and sometime collapsed very early. We gradually increased the margin from $0.0$ to $0.4$ and observed that RMCosGAN achieved the best FID with margin $m$ set to $0.15$ on both datasets. This demonstrates the importance of the margin $m$ and its setting on RMCosGAN performance. Thus, we set the margin to $0.15$ for all experiments.

\subsubsection{Impact of the number of generated images on FID.} To investigate the impact of the number of generated images on FID, we generated different sets of images and computed the FID value for each set. In Fig.~\ref{fig:impact_cifar10} and Fig.~\ref{fig:impact_mnist}, we present the experimental results when training RMCosGAN on CIFAR-10 and MNIST, respectively. We observed that FID decreases along with the increase in the number of images generated. This shows that the higher the number of images generated, the higher the similarity among them. This also shows that the RMCosGAN has been well trained to generate images with similar features learned from the training datasets.

\subsubsection{Huge performance variance on the MNIST dataset.}
\begin{figure}[t]
    \centering
    \includegraphics[width=\textwidth]{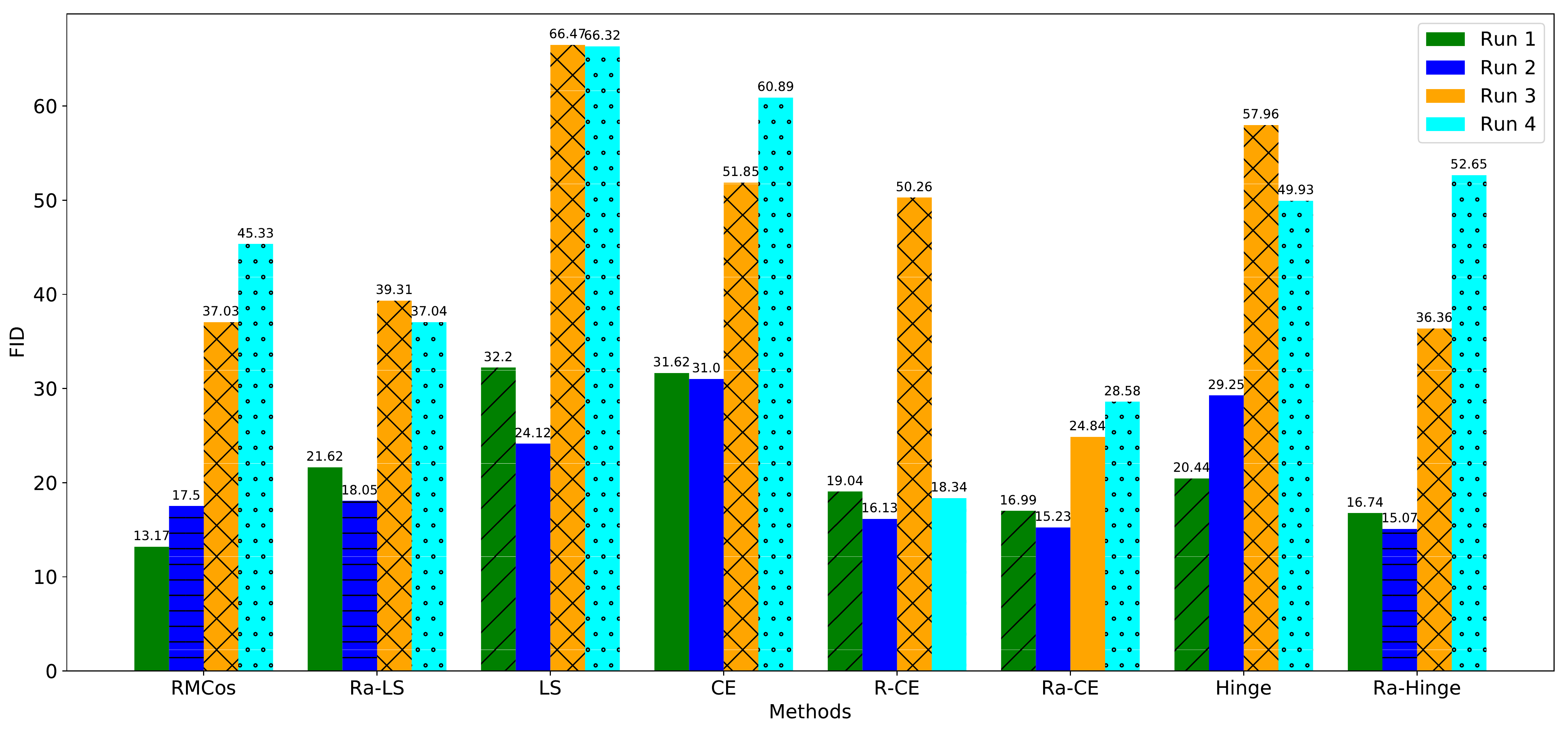}
    \caption{Best FID of all methods on the MNIST dataset after four different runs.}
    \label{fig:huge_variance_on_mnist}
\end{figure}

As presented in the previous section, we have used two workstations for all the experiments. During the training of RMCosGAN on MNIST, we observed a huge variance on FID after each run. In Fig.~\ref{fig:huge_variance_on_mnist}, we present the obtained results. On the one hand, it is important to emphasize that FID is measured based on the generated images and the pre-trained-on-ImageNet Inception classifier. However, the distributions of MNIST and ImageNet do not line up identically. On the other hand, the difference in computing hardware architectures could also contribute to this variance. Thus, a small change in network weights would remarkably affect the performance. 

\begin{table}[t]
\centering
\caption{FID of RMCosGAN achieved with all the  datasets at 100K training epochs}
\label{tab:theorem1_FID}
\begin{tabular}{l@{\hskip 0.2in} r@{\hskip 0.2in} r@{\hskip 0.2in} r@{\hskip 0.2in} r}
\hline
\textbf{m} & \textbf{CIFAR-10} & \textbf{MNIST} & \textbf{STL-10} & \textbf{CAT}\\
\hline\hline
$m<0$ & $37.35\pm0.15$ & $17.92\pm0.07$ & $56.89\pm0.39$ & $10.36\pm0.14$\\ 
\hline
$m>0$ & $\mathbf{31.34}\pm\mathbf{0.13}$ & $\mathbf{13.17}\pm\mathbf{0.05}$ & $\mathbf{52.16}\pm\mathbf{0.35}$ &  $\mathbf{9.48}\pm\mathbf{0.11}$\\ 
\hline
\end{tabular}
\end{table}

\begin{table}[t]
\centering
\caption{IS of RMCosGAN achieved with all the  datasets at 100K training epochs}
\label{tab:theorem1_FID}
\begin{tabular}{l@{\hskip 0.2in} r@{\hskip 0.2in} r@{\hskip 0.2in} r@{\hskip 0.2in} r}
\hline
\textbf{m} & \textbf{CIFAR-10} & \textbf{MNIST} & \textbf{STL-10} & \textbf{CAT}\\
\hline\hline
$m<0$ & $6.41\pm0.07$ & $2.02\pm0.03$ & $6.34\pm0.05$ & $3.22\pm0.04$\\ 
\hline
$m > 0$ & $\mathbf{7.08}\pm\mathbf{0.06}$ & $2.18\pm0.02$ & $\mathbf{6.86}\pm\mathbf{0.03}$ & $3.58\pm 0.02$ \\ 
\hline
\end{tabular}
\end{table}

\section{Conclusions}
\label{sec:conclusion}

In this paper, we carried out an empirical study on the instability of GANs and the impact of loss functions on GAN performance. We adopted the standard GAN architecture and implemented multiple loss functions for comparison purposes. We developed a novel loss function (RMCosGAN) that takes advantage of the relativistic discriminator and incorporates a margin into cosine space. While the relativistic discriminator improves the stability of GAN training, the margin cosine loss can enhance the discrimination between real and generated samples. We carried out extensive experiments on four benchmark datasets. We also carried out an ablation study to evaluate the impact of different parameters on the performance of GANs. The experimental results show that the proposed loss function (RMCosGAN) outperforms the existing loss functions, thus improving the quality of images generated. The results also show that using RMCosGAN achieves a stable performance throughout all the datasets regardless of their distribution. It is worth mentioning that the performance comparison of different GAN models is always challenging due to their instability and hyper-parameters sensitivity. While we used a standard GAN architecture in this work, it would be interesting to explore more complex architectures. Furthermore, it would also be interesting to experiment with different neural networks for discriminators and generators such as recurrent neural networks in the domain of text generation. 

\bibliographystyle{splncs04}
\bibliography{iconip2020}

\begin{subappendices}
\label{appendix}

\section{Randomly-generated Images}

\begin{figure}[!h]
    \centering
    \includegraphics[width=1.0\textwidth]{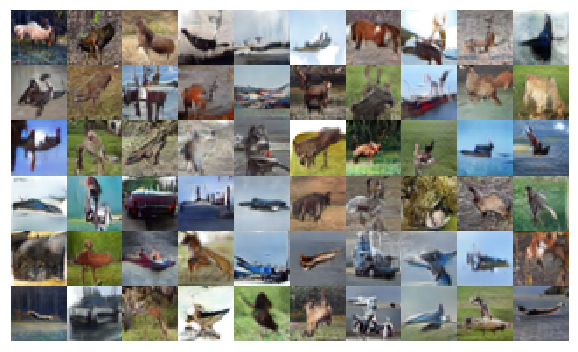}
    \caption{60 randomly-generated images using our proposed RMCosGAN at $\text{FID}=31.34$ trained on CIFAR-10 dataset.}
    \label{fig:images_cifar}
    \vspace{-5ex}
\end{figure}
 
\begin{figure}[!h]
    \centering
    \includegraphics[width=1.0\textwidth]{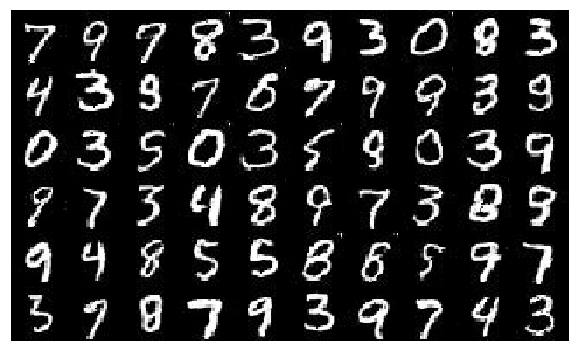}
    \caption{60 randomly-generated images using our proposed RMCosGAN at $\text{FID}=13.17$ trained on MNIST dataset.}
    \label{fig:images_mnist}
    \vspace{-5ex}
\end{figure}

\begin{figure}[!h]
    \centering
    \includegraphics[width=1.0\textwidth]{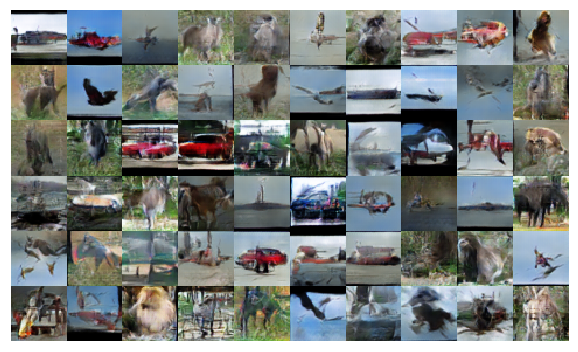}
    \caption{60 randomly-generated images using our proposed RMCosGAN $\text{FID}=52.16$ trained on STL-10 dataset.}
    \label{fig:images_stl10}
    \vspace{-5ex}
\end{figure}
 
\begin{figure}[!h]
    \centering
    \includegraphics[width=1.0\textwidth]{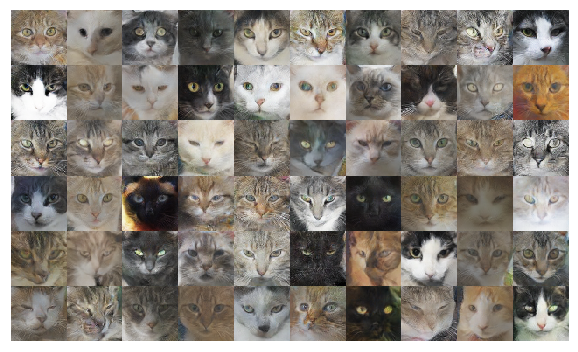}
    \caption{60 randomly-generated images using our proposed RMCosGAN at $\text{FID}=9.48$ trained on CAT dataset.}
    \label{fig:images_cat}
    \vspace{-5ex}
\end{figure}

\end{subappendices}

\end{document}